\documentclass[10pt,conference]{IEEEtran}
\IEEEoverridecommandlockouts
\usepackage{amsmath,amssymb,amsfonts}
\usepackage{algorithmic}
\usepackage{graphicx}
\usepackage{dblfloatfix}
\usepackage{textcomp}
\usepackage{xcolor}
\usepackage{float}
\usepackage{multirow}
\usepackage{footnote}
\usepackage{mathtools}
\usepackage[font=small,skip=0pt]{caption}
\usepackage{mathtools}

\usepackage{booktabs}
\usepackage{amsthm}

\usepackage[nolist]{acronym}

\usepackage[
	pdfborder={0 0 0},
	colorlinks=false,%
	urlcolor=black,%
	linkcolor=black,%
	citecolor=black,%
	filecolor=black,%
	breaklinks,%
	]{hyperref}%
	
\usepackage[%
	backend=biber,%
	isbn=false,%
	hyperref=true,%
	maxbibnames=99,%
	sorting=none,%
	natbib=true,%
	language=english,%
	]{biblatex}%

\usepackage{setspace}

\usepackage[letterpaper,
            top=0.75in,    
            bottom=1in,    
            left=0.65in,   
            right=0.65in]{geometry}

\newtheorem{theorem}{Theorem}[section]   
\newtheorem{proposition}[theorem]{Proposition}       
\theoremstyle{definition}

\DeclareFieldFormat{sentencecase}{\csname bbx@colon@search\endcsname#1} 

\def\BibTeX{{\rm B\kern-.05em{\sc i\kern-.025em b}\kern-.08em
    T\kern-.1667em\lower.7ex\hbox{E}\kern-.125emX}}

\begin{document} 

\title{ Knowledge vs. Experience: Asymptotic Limits of Impatience in Edge Tenants
}

\author{
	\IEEEauthorblockN{Anthony~Kiggundu\IEEEauthorrefmark{1}\IEEEauthorrefmark{2},~Bin~Han\IEEEauthorrefmark{2}
    ~and~Hans~D.~Schotten\IEEEauthorrefmark{1}\IEEEauthorrefmark{2}}
	\IEEEauthorblockA{
		\IEEEauthorrefmark{1}German Research Center for Artificial Intelligence (DFKI), Germany\\
		\IEEEauthorrefmark{2}RPTU University of Kaiserslautern-Landau, Germany\\
	}
}

\maketitle

\begin{abstract}
We study how two information feeds, a closed-form Markov estimator of residual sojourn and an online trained actor–critic, affect reneging and jockeying in a dual M/M/1 system. Analytically, for unequal service rates and total-time patience, we show that total wait grows linearly so abandonment is inevitable and the probability of a successful jockey vanishes as the backlog $\to\infty$. Furthermore, under a mild sub-linear error condition both information models yield the same asymptotic limits (robustness). We empirically validate these limits and quantify finite backlog differences. Our findings show that learned and analytic feeds produce different delays, reneging rates and transient jockeying behavior at practical sizes, but converge to the same asymptotic outcome implied by our theory. The results characterize when value-of-information matters (finite regimes) and when it does not (asymptotics), informing lightweight telemetry and decision-logic design for low-cost, jockeying-aware systems.


\end{abstract}


\section{Introduction} 
\label{sec:introduction}
Classical queueing theory supplies the probabilistic foundation for analyzing sojourn times and customer impatience—specifically reneging (abandoning) and balking (refusing entry)—which are fundamental to service system design \cite{Kleinrock1976,HarcholBalter2013}. This foundational work is complemented by a rich literature on jockeying(switching queues) and threshold control policies formalized by \ac{MDP}, which characterize optimal switching boundaries in small or homogeneous systems \cite{Takagi1991}. 
Despite this theoretical basis, research on active queue management (AQM) \cite{Claypool} and the value of state information often evaluates routing and congestion control heuristics without rigorously linking the scaling of estimator error to the robustness of the decision policies \cite{kiggundu2025nfvsdn,Ibrahim}. Simultaneously, the deployment of technologies like \ac{MEC}, slice orchestration, and \ac{NWDAF} makes per-request migration operationally feasible. This technological shift motivates dynamic behavior like reneging and jockeying but creates new trade-offs between control plane overhead and the potential migration benefit \cite{etsi_ts122261_v16}. These incentives for dynamic behavior are amplified in modern edge/MEC systems, which expose tenants to fluctuating queueing delays and compel them to make real-time decisions about reneging or jockeying \cite{kiggundu2024,mahabhashyam2005queues}.
\subsection{Motivation}
These individual decisions create significant externalities. A tenant optimizing their personal utility necessarily ignores the marginal effect of their departure or switch on others, such as changes in queue lengths and waiting times. For the system provider, higher reneging leads to fewer completions and thus lower revenue, though it may benefit remaining tenants by reducing queue lengths. Similarly, while jockeying can lead to better service for the switching tenant, it introduces new system inefficiencies like increased control overhead, more variability, and potentially longer waits for others. Managing these trade-offs via for example pricing, quotas, coordinated telemetry or admission/slicing controls, is therefore central to stable, efficient queue management.
While prior work has studied this impatience in queueing systems \cite{Kiggundu2024ChroniclesOJ}, proposing analytic estimators or heuristic migration rules, there is a limited formal treatment of two critical aspects: (i) how on-device learned policies compare to closed-form queueing estimators in the same stochastic setting, and (ii) the interplay between estimator fidelity, telemetry cost, and the asymptotic queue behavior under a total-time patience model. Moreover, existing applications of Reinforcement Learning (actor–critic) to networking problems seldom compare empirical results against closed-form queueing predictors under rigorous asymptotics \cite{Konda2000,Bhatnagar2009}.
\subsection{Contribution of this Paper}
In this paper, we address these gaps by comparing two practical information feeds used to guide reneging and jockeying decisions: (i) a closed-form Markov chain estimator of residual sojourn time and (ii) an online-trained actor–critic policy that encodes past experience. Analytically, we demonstrate that under the realistic total-time patience model and a mild regularity condition on estimator error:
\begin{itemize}
  \item \textbf{Asymptotic analysis (unequal rates).} The remaining service time grows linearly with the number of requests in heterogeneous queues. Consequently, for any fixed patience threshold, the unconditional probability of abandonment tends to one, and the probability that a jockeyed request finishes before its deadline vanishes as the queue lengths $\to \infty$.
  \item \textbf{Estimator robustness.} We prove a robustness result showing that, under a natural sublinear-error condition, both the Markov and learned predictors produce the same asymptotic behavior, even if they differ significantly at practical finite backlogs.
  \item \textbf{Finite-queue occupancy characterization.} We provide empirical results and metrics showing how learned and analytic feeds differ at practical queue lengths, and identify parameter regimes where the extra telemetry or learning complexity gives measurable value.
  \item \textbf{Design guidance.} Based on theory and experiments, these findings inform lightweight telemetry and decision-logic choices for low-cost, jockeying-aware systems such as inter-slice switching in \ac{5G}/\ac{6G} and edge offload platforms.
\end{itemize}
In the next section we characterize for impatience emerging from the subscription to the two information feeds. Section \ref{sec:results} documents the results comparing the influence of the information models on the behavior of the tenant. We conclude with some insights into pending work in Section \ref{sec:conlusion}. 

\section{System Modeling} 
\label{sec:System Model} 
\subsection{Setup and Assumptions}
Our setup assumes two \ac{M/M/1} queues $i,j$ with Poisson distributed admissions at rate $\lambda$ and exponential service rates $\mu_{i},\mu_{j}$. The service discipline is \ac{FCFS} and the total arrival rate $\lambda=\lambda_{i}+\lambda_{j}$. 
Each request has a random patience \(T\) measured from the instant of entry into the system. If a request spends time \(t_0\) before switching, its remaining patience after switching is \(T-t_0\). A request that abandons the system  takes $T_{local}$ time to get processed on the local device. Heterogeneity (relative to total arrival rate): we define $\delta_{\lambda}$, a random parameter $-\lambda<\delta_{\lambda}<\lambda$ such that     $\mu_i=\frac{\lambda_{i}+\delta_{\lambda}}{2}, \mu_j=\frac{\lambda_{j}-\delta_{\lambda}}{2}.$ A request that jockeys from queue $i$ to $j$ has $0\leq \lambda_{tar}\leq \lambda_{j}$ that join ahead of it at switching time($\lambda_{tar}$ arrivals infront of jockey).    

\subsection{Model of the Knowledge: Markovian Waiting time}
For a pending request that has $k$ other requests ahead in the current queue $i$, the total remaining time until servicing is expressed for using \eqref{eqn:waiting_times}
\begin{equation}
      \small
       W_i(k)=\sum_{t=1}^k X_{i,t},\qquad X_{i,t}\overset{\mathrm{iid}}{\sim}\mathrm{Exp}(\mu_i).
    \label{eqn:waiting_times}
\end{equation}  
Here, $W_i(k)$ is \textit{Erlang}$(k,\mu_{i})$, its expectation and respective density function is defined using \eqref{eqn:expwait}.
\begin{equation} 
    \small
        \mathbb{E}[W_i(k)] = \frac{k}{\mu_i}, \quad 
        f_{W_i(k)}(t)=\frac{\mu_i^k\,t^{\,k-1}e^{-\mu_i t}}{(k-1)!},\; t\ge0. 
    \label{eqn:expwait}
\end{equation}
\subsubsection{Reneging under Markovian knowledge} 
From \eqref{eqn:waiting_times}, if the remaining time to service completion is more than the patience time $T<T_{local}$, then it makes more sense to renege from the queue. For an $M/M/2$ setup, the buffered request at position $k$ in queue $i$ will prefer local processing with probability defined by \eqref{eqn:renege_probab}. 
\begin{equation}
    \small
        P_{\rm reneg}\bigl\lvert k\bigr.
        =\Pr\{W_{i} > T\,\mid\,k\}
        =1 - \Pr\{W_{i} \le T\mid k\} 
    \label{eqn:renege_probab} 
\end{equation}
However, if the request chooses to stay at decision time with elapsed $t_0$, the probability it reneges before service completion (i.e., fails to complete before patience expires) is defined by \eqref{eqn:bad_renege_prob}.
\begin{equation}
    \small
    \begin{split}
        \Pr_{\text{reneg\_fail}}\bigl(t_0 + W_i(k) \le T\bigr)
         = \Pr\bigl(W_i(k) \le T-t_0\bigr) \\
        = F_{W_i(k)}(T -t_0).
    \end{split}
     \label{eqn:bad_renege_prob}
\end{equation}

\subsubsection{Jockeying under Markovian knowledge} 
\paragraph{\textbf{Pure-death case}: arrivals $\lambda_{tar}=0$ join ahead of jockey}
The Pure-Death Process is a special case of the birth-death process where the birth rate is identically zero for all states (i.e., $\lambda_{i\neq j}\!=\!\lambda_{\text{tar}}=0$). This means the system size can only decrease over time, leading the process to eventually become absorbed in state $0$, unless the state space is infinite and the process is explosive
If $\lambda_{\rm tar}=0$ while the jockey is in transit, then the chain only jumps downward. That is, every customer present departs at rate $\mu$ independently (so the total departure rate in state $n$ is $n\mu$). This has the implication that each of the $n$ initial customers survives until time $t$ independently with probability $e^{-\mu t}$. Hence the number of requests $k$ remaining at time $t$ is binomial and defined by \eqref{eqn:binomal}.
\begin{equation}
    p_{n\to k}(t)
    = \binom{n}{k}(e^{-\mu t})^{k}\bigl(1-e^{-\mu t}\bigr)^{n-k},
    \qquad 0\le k\le n.
    \label{eqn:binomal}
\end{equation}
\textit{where $n-k$ is the number of deaths.} 

\noindent For the $M/M/2$ setup in the state $n$ therefore, the death rates repeat $\mu_n=2\mu$ for $n\ge2$. The time to reduce from state $n$ down to the first initial state ($\{0,\! \dots,\! c\!-\!1\}$, $c$ \textit{queues}) is simply a sum of independent exponential means $ \sum_{j=c}^{n} \frac{1}{\mu_j}$ (with the convention that the sum is zero if $n<c$). 
Therefore the expected time $T_{\rm jockey}$ to go from $n$ down to being served is
\begin{equation}
    T_{\rm jockey} = \sum_{j=2}^{n}\frac{1}{2\mu} = \frac{n-1}{2\mu},\qquad n\ge2.
\end{equation}
So in the pure-death approximation, the waiting time until service start is linear in (n) with slope $1/(2\mu)$ for states $n\ge2$. Therefore, the unconditional expected time for the jockeyed request is 
\begin{equation}
    \mathbb{E}[T_{\rm jockey}] = 
    \sum_k p_{n\to k}(t) T_{\rm jockey} 
\end{equation}
\paragraph{\textbf{Birth-death case:} $\lambda_{tar}>0$ join ahead of jockey}
A birth--death process is the general case where transitions change the current state of a queue $i$ given arrivals (births) to that queue at a rate $\lambda_{i}$ ($\lambda_{tar}<\lambda_{i}$ join ahead of jockeyed request) and departures (deaths) at a rate $\mu_{i}$.
For the $M/M/2$ setup, the birth–death \ac{CTMC} is composed of departure rates $\mu_n=\min(n,c)\mu$ ($c$ queues) when in state $n$ and a portion of arrivals $0<\lambda_{tar}<\lambda_{i\neq j}$. The CTMC generator matrix has off-diagonals $q_{n,n+1}=\lambda_{\mathrm{tar}}$ and $q_{n,n-1}=\mu_n$. For this chain, we characterize for the transient \ac{PMF} of the number of requests in the alternative queue at time $t$ from uniformization principles (a Poisson-mixture of powers of a simple discrete-time Markov matrix $P$). That is, we pick a uniformization rate $q = \lambda_{\mathrm{tar}} + 2\mu$ (or any $q\ge \lambda_{\mathrm{tar}}+\max_n\mu_n)$ $q\ge \lambda_{\mathrm{tar}}+\max_n\mu_n$ and define the one-step tri-diagonal stochastic matrix $P$ with the only non-zero entries using \eqref{eqn:trientries}. 
\begin{equation}
    \small
    \begin{aligned}
        &P_{n,n+1} = \frac{\lambda_{\mathrm{tar}}}{q},\qquad n\ge0, \\ 
        &P_{n,n-1} = \frac{\mu_n}{q},\qquad n\ge1,\\ 
        &P_{n,n} = 1-\frac{\lambda_{\mathrm{tar}}+\mu_n}{q},
    \end{aligned}
    \label{eqn:trientries}
\end{equation}
We let $X(t)$ denote the state of the CTMC with state space $\left\{0,1,2,\! \ldots \right\}$, then $\pi^{(0)} = \Pr\left\{X(0)=k\right\}$, $k=0,1,2,\! \ldots$. And the initial distribution definitive of how the queue length (or state) is distributed is the row vector $\pi^{(0)} = \left[\!\pi_{0}^{(0)}\!,\! \pi_{1}^{(0)}\!,\! \pi_{2}^{(0)}\!,\! \ldots \!\right]$.
If this initial distribution is unknown, we define it using \ref{eqn:init_distr}. 
\begin{equation}
    \small
    \pi^{(0)}= \Biggl(1 + 2\rho + \dfrac{2\rho^2}{1-\rho}\Biggr)^{-1}, \quad \rho = \frac{\lambda}{c \mu}
    \label{eqn:init_distr}
\end{equation} 
Then denoting as $N(t)$ the number of requests in the target queue at time $t$, \eqref{eqn:mass_funct} defines for the PMF for $k$ customers as the $k^{th}$ component.
\begin{equation}
    \small
    \Pr\{N(t)=k\} = \sum_{m=0}^{\infty} e^{-q t}\frac{(q t)^m}{m!}\bigl(\boldsymbol{\pi}^{(0)}P^m\bigr)_k.
    \label{eqn:mass_funct}
\end{equation} 
Let $m_{k}$ be the expected time until the tagged jockey reaches service starting from state $k$ ($k<c$  customers in the target queue). Then $m_{0}=0$ and for each state $n \geq 1$, the standard first-step conditioning (birth–death hitting time equation) gives the recurrence $(\lambda_{\mathrm{tar}}+\mu_n)m_n - \lambda_{\mathrm{tar}}m_{n+1} - \mu_n m_{n-1} = 1$. 
For the $M/M/2$ system, $\mu_n=2\mu$ for $n\ge2$, implying that for $n\ge2$, this recurrence is re-defined as $(\lambda_{\mathrm{tar}}+2\mu)m_n - \lambda_{\mathrm{tar}}m_{n+1} - 2\mu m_{n-1} = 1$ and the solution for this recurrence is affine linear in $n$. 
In equilibrium ($\lambda_{\mathrm{tar}}<2\mu$) therefore, \eqref{eqn:remain_time} defines for the  closed form solution of the remaining waiting time for the jockey landing at position $k$. 
\begin{equation}
    \small
    T_{\rm jockey} = \dfrac{k-1}{2\mu-\lambda_{\mathrm{tar}},}, \quad k\ge2.
    \label{eqn:remain_time}
\end{equation}
Then the expected jockey time until service start is 
\begin{equation}
    \small
    \mathbb{E}[T_{\rm jockey}]
    =
    \sum_{k=0}^{\infty} (\Pr\{N(t)=k\})T_{\rm jockey} 
\end{equation}
Numerically, we compute $\Pr{N(t)=k}$ up to $k\le N_{\max}$ using uniformization truncation such that this tail is truncated when its mass is negligible.
From total probability principles the impatient tenant therefore switches to the alternative queue with probability defined as a mixed integral by \eqref{eqn:decide_jockey}.
\begin{equation}
    \small
    \begin{split}
        \Pr\bigl(T_{\rm jockey} < W_i(k)\mid N(t)=k) 
        = \int_0^\infty f_{W_i(k)}(u) \Pr\bigl(T_{\rm jockey} < u\bigr) du \\
        = \int_0^\infty f_{W_i(k)}(u)\Bigl(\int_0^{u} g_k(t)dt\Bigr) du. 
    \end{split}    
    \label{eqn:decide_jockey}
\end{equation}
\textit{where $g_k(t)$ is the distribution function definitive of $T_{jockey}$ given $N(t)=k$.} 

\noindent For the selfish and impatience tenant, the probability that after jockeying that tenant reneges (fails to finish before patience) is characterized by \eqref{eqn:failed_jockey}
\begin{equation}
    \small
      \Pr_{\text{switch\_fail}} = \sum_k \Pr\{N(t)=k\}\int_{T-t_0}^\infty g_k(u)\,du. 
    \label{eqn:failed_jockey}  
\end{equation}
\textit{where $g_k(u)$ is the density function of the remaining time when joining at position $k$}. 

\noindent On the other hand, the probability of a successful jockey (switch and finish before patience) is defined using \eqref{eqn:succ_jock}.
\begin{equation}
    \small
    \Pr_{\text{switch\_succ.}} = 1 - \Pr_{\text{switch\_fail}}
    = \sum_k \Pr\{N(t)=k\}\int_0^{T-t_0} g_k(u)\,du. 
    \label{eqn:succ_jock} 
\end{equation}
\subsection{Model of the Experience: Actor-Critic}
\noindent We let $s=(k_{i},k_{j},\mu_{i},\mu_{j},T)$ denote the state of queues, such that $k_{i}$ is the position of the customer in its current queue $i$ ($\mu_{i}$), $k_{j}$ is the corresponding position in an alternative queue $j$ ($\mu_{j}$) and $T$ the impatience. 
Tenants then subscribe to an actor–critic information source and at each step observes the queue state, samples an action $a\in{0,1}$ (renege or jockey respectively) from its policy, and receives a reward and next state. To recalibrate the sensitivity of the reward to the difference in time, we define $\tau>0$ and $\delta>0$ as non-negative scaling factors to represent the utility cost of time or preference for a quick decision.
A tenant in the state $s$ will prefer taking an action $a$ with probability defined $\pi_{\theta}(a|s)$. 
The reward function for our actor-critic is characterized by the behavioral component (jockey or renege) and the respective learned probability for the specific behavior. Equations \eqref{eqn:new_reward_funct} and \eqref{eqn:new_jockey_funct} define for these rewards given the reneging and jockeying actions respectively.
\begin{equation}
    \mathbf{R}(s,a=0) = \pi_{\theta}(0|s). \frac{1}{1+e^{\bigg [- \bigg (\frac{k_{i}}{\mu_{i}}- T \bigg) \bigg ]}} 
    \label{eqn:new_reward_funct}
\end{equation} 
\begin{equation}
    \mathbf{R}(s,a=1) = \pi_{\theta}(1|s). \frac{1}{1+e^{\bigg [- \bigg (\frac{k_{i}}{\mu_{i}} - \frac{k_{j}}{\mu_{j}}  \bigg) \bigg ]}} 
    \label{eqn:new_jockey_funct} 
\end{equation}
\subsubsection{The Value Function and the Optimal Policy}
The discounted return is defined for by \eqref{eqn:discounted} and is fundamental for formulating \eqref{eqn:bellman} as the state value function. 
\begin{equation}
   G_{t} = \sum^{\infty}_{k=0} \gamma^{k}r_{{t+k}}
    \label{eqn:discounted}
\end{equation}
\textit{where $\gamma \in (0,1)$ denotes the discount factor that regulates the rewarding process given the uncertainties in the subsequent environment states.}
\begin{equation}
   V^{*}(s) = \max_{\pi} \mathbb{E} \bigg [ G_{t} | s_{t}=s\bigg]
    \label{eqn:bellman}
\end{equation}
Adopting Bellman's conditions for optimality then \eqref{eqn:bellman} evolves into \eqref{eqn:bellmanfull}.
\begin{equation}
    V^{*}(s) = \max_{a\in\{0,1\}} \bigg [ \mathbf{R}(s,a) + \gamma\sum_{s^{'}} p(s^{'}|s,a)V^{*}(s^{'}) \bigg ]
    \label{eqn:bellmanfull}
\end{equation}

For typical Markov Decision Processes, the maximum expected return attained when tenants act in a given state is characterized for using the action-value function $Q^{*}(s,a)$. The Bellman equivalent for this \textit{Q-function} in our setting is:
\begin{equation}
   Q^{*}(s,a) = \mathrm{R}(s,a) + \gamma \mathbb{E}_{s\sim {p(.|s,a)}} \bigg[ \max_{a^{'}\in \mathcal{A}} Q^{*} (s^{'}, a^{'}) \bigg]
    \label{eqn:bellQ}
\end{equation} 
\textit{where $R(s,a), p(s^{'}|s,a)$ denote instantaneous reward from an action $a$ in state $s$ and the subsequent probability of ending up in the next possible state $s^{'}$ given the action.}

\noindent The resultant action-value function therefore incorporates this instantaneous reward plus any future reward, such that for any jockeying or a reneging activity, \eqref{eqn:bellQ} is redefined for each of these activities into:
\begin{equation}
  \small
  \begin{split}    
      Q^{*}(s,0) = \mathrm{R}(s,a=0) + \gamma \mathbb{E}_{s\sim {p(.|s,a)}} \bigg[ max\{ Q^{*}(s^{'}, 0), Q^{*}(s^{'},1)\} \bigg] \\
      Q^{*}(s,1) = \mathrm{R}(s,a=1) + \gamma \mathbb{E}_{s\sim {p(.|s,a)}} \bigg[ max\{ Q^{*}(s^{'}, 0), Q^{*}(s^{'},1)\} \bigg]      
  \end{split}    
    \label{eqn:qfunct}
\end{equation}

Substituting the reward functions \eqref{eqn:new_reward_funct} and \eqref{eqn:new_jockey_funct} corresponding to the respective behavior yields the action-value function formulations \eqref{eqn:actvalue_rene} for the reneging activity:
\begin{align}
  &\begin{aligned}
    \mathllap{ Q^{*}(s,0)} &=  \pi_{\theta}(0|s) \cdot \frac{1}{1+e^{\bigg [- \bigg (\frac{k_{i}}{\mu_{i}}- T \bigg) \bigg ]}} \\ 
      &\qquad + \gamma \mathbb{E}_{s\sim {p(.|s,0)}} \bigg[  \max_{a^{'}\in \{ 0,1\}} Q^{*} (s^{'}, a^{'})\} \bigg]
  \end{aligned} 
  \label{eqn:actvalue_rene}
\end{align}

And \eqref{eqn:actvalue_jock} for the jockeying activity as:
\begin{align}
     &\begin{aligned}
    \mathllap{ Q^{*}(s,1)} &=\pi_{\theta}(1|s) \cdot \frac{1}{1+e^{\bigg [- \bigg (\frac{k_{i}}{\mu_{i}} - \frac{k_{j}}{\mu_{j}}  \bigg) \bigg ]}}\\ 
      &\qquad +  \gamma \mathbb{E}_{s\sim {p(.|s,1)}} \bigg[  \max_{a^{'}\in \{ 0,1\}} Q^{*} (s^{'}, a^{'})\} \bigg]
  \end{aligned} 
  \label{eqn:actvalue_jock}
\end{align}

And the optimal policy is then given by \eqref{eqn:optimal}
\begin{equation}
    \pi^{*}(s) = \arg \max_{a \in \{0,1\}} Q^{*}(s,a)
    \label{eqn:optimal}
\end{equation}
From the characterization of these information sources and the hyperactivity arising from the tenants impatience, we draw the following proposition:

\begin{proposition}[Asymptotic behavior of jockeying and reneging under total time patience]
\label{prop:jockey-asymptotics}
Let queues 1 and 2 be independent \ac{M/M/1} with service rates $\mu_{1},\mu_{2}$. Assume an arriving customer faces $n$ jobs in queue 1 and changing count of queued requests $m$ in queue 2, and draws a finite patience $T$ at entry which is consumed continuously (total time patience). If $\mu_{1}\neq \mu_{2}$, then as $n\to\infty$: (i) the unconditional probability of reneging satisfies $\Pr(\text{renege})\to 1$, and (ii) the probability of a successful jockey (switch and complete service before $T$) satisfies $\Pr(\text{successful jockey})\to 0$. 
\end{proposition}
Reneging $\to 1$: Because the first queue’s wait grows like $n/\mu_1$ and $T$ is fixed, the initial slow delay dominates — switching cannot “erase” the time already spent. So abandonment is almost certain.
Successful jockeying $\to 0$: Any event that results in completion of service before $T$ is contained in $\{W_{\rm total}\le T\}$; that rare event’s probability $\to 0$. Thus even if customers attempt to switch, those attempts mostly do not lead to successful service but in abandonment.

\textit{Proof 1: See Appendix}
\begin{theorem}[Robustness to information models — unequal rates]
\label{thm:robustness_unequal}
Let queues $1,2$ be independent $M/M/1$ with service rates $\mu_1,\mu_2$ and assume $\mu_1\neq\mu_2$.
An arrival finds $n$ customers ahead in queue~1 and (if switching) a fixed finite number of queued requests $m$ in queue~2.
Let $W_i(k)=\sum_{t=1}^k X_{i,t}$ with $X_{i,t}\stackrel{\mathrm{iid}}{\sim}\mathrm{Exp}(\mu_i)$ be the remaining work when $k$ jobs are ahead.
Decisions to jockey are based on estimates $\widehat W_i(\cdot)$ provided by an information model (Markov predictor, NN, etc.).
Assume the information models satisfy the \emph{sublinear error} condition
\begin{equation}
    \small
    \frac{|\widehat W_i(n)-W_i(n)|}{n}\xrightarrow{\mathbb{P}}0\qquad (n\to\infty),
\end{equation}
for $i=1,2$.

\noindent Under total time patience (The total waiting time $T$ is not reset when a request jockeys), as $n\to\infty$, independently of the information model:
\begin{enumerate}
  \item  the unconditional abandonment probability satisfies   \[
    \Pr\{\text{reneging}\}=\Pr\{W_{\rm total}>T\}\xrightarrow{n\to\infty}1,
  \]
  \item and consequently the probability of a \emph{successful} jockey (switch \emph{and} complete service before $T$) tends to $0$.
\end{enumerate} 
\end{theorem}

\textit{Proof 2: See Appendix}

\section{Simulation Results}
\label{sec:results}
The arrival rates in our setup were randomized such that at each iteration this rate was sampled from the set $\lambda\in[3,5,7,9,11,13,15]$. 
Our empirical experiment was characterized by \textit{100 episodes}, each episode iteration numbered \textit{100 epochs}. The actor-critic is a two-layer network for both policy and value estimation, each layer with \textit{128 hidden units} and \textit{ReLU activation}. We embedded a \textit{softmax} function on the actor for action selection and a \textit{scalar} output on the critic for value estimation. The actor-critic was instantiated with a learning rate of $0.001$ and an \textit{Adam optimizer} with cross entropy.

\begin{figure}
   \centering
     \includegraphics[width=0.48\textwidth]{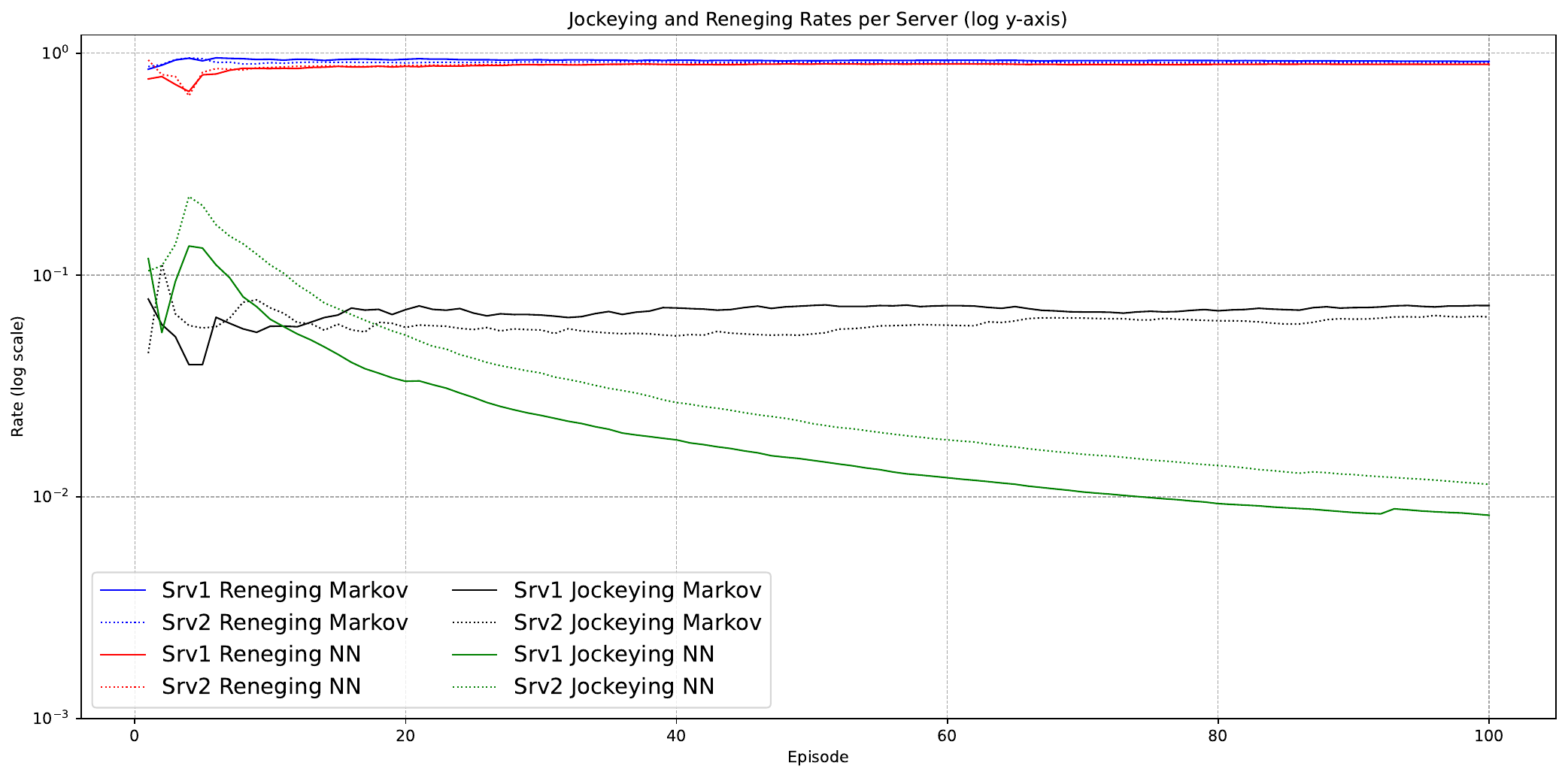} 
     \caption{\footnotesize Under total-time patience, as queue lengths grow large the time already spent waiting in the original queue overwhelms any finite patience budget. Therefore almost every arrival abandons before being served, and the fraction of arrivals who actually switch and get served collapses to zero. }
  \label{fig:rate_jock_reneg}
\end{figure}

\begin{figure}
  \centering
    \includegraphics[width=0.47\textwidth]{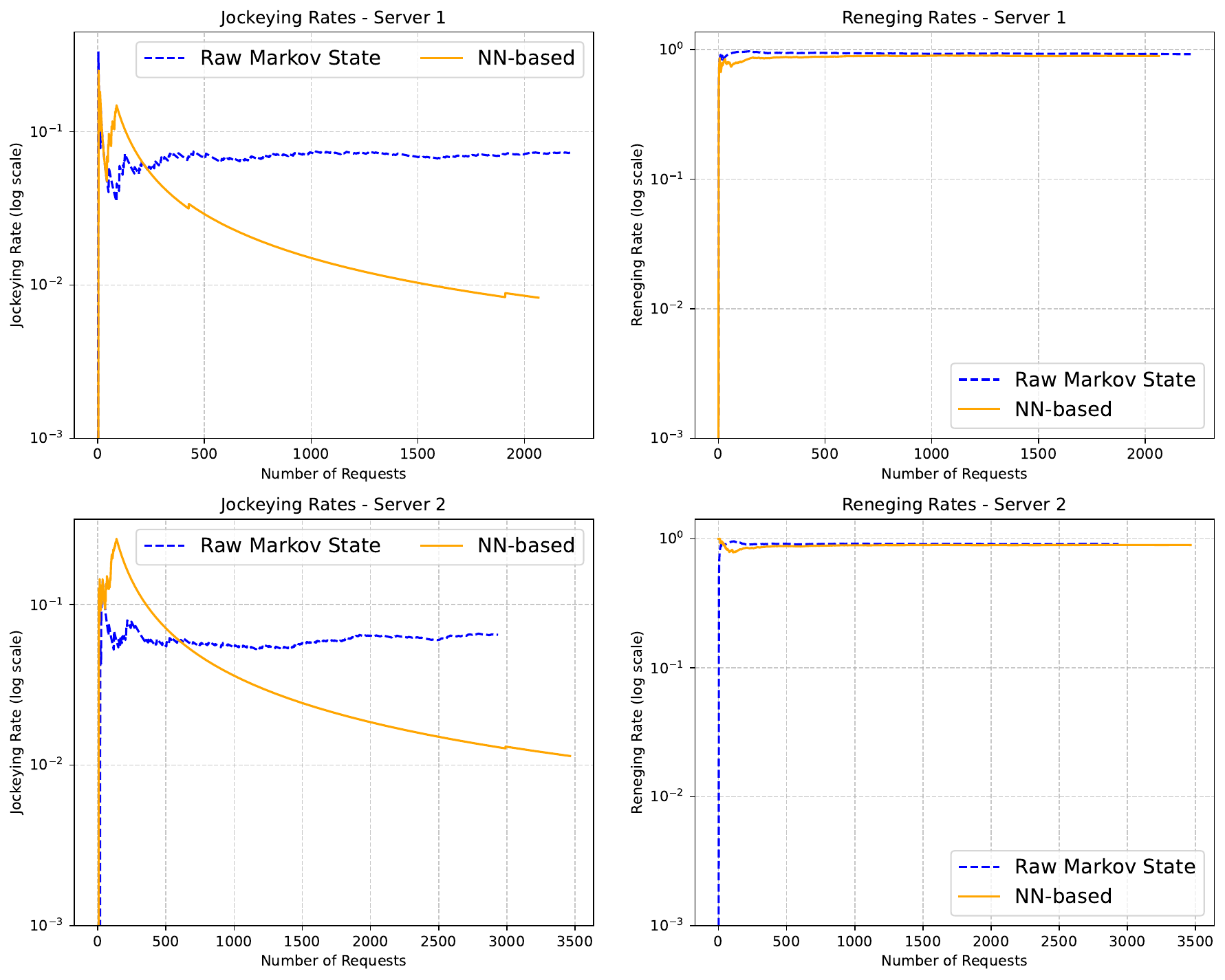} 
    \caption{\footnotesize As the system grows in size, jockeying becomes negligible and reneging stabilizes at a high rate for both servers, regardless of whether decisions are guided by raw Markov state or NN-based information.}
  \label{fig:rates_info_src}
\end{figure}

\begin{figure} 
    \centering
        \includegraphics[width=0.47\textwidth]{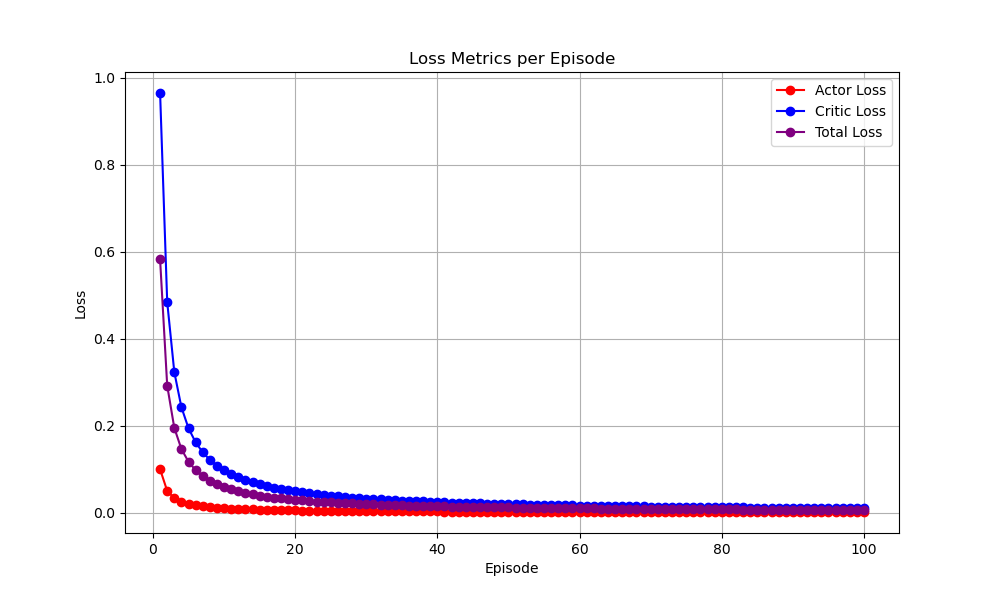}
    \caption{\footnotesize Loss metrics for the actor and critic networks decrease rapidly over episodes, indicating effective learning and convergence of the actor-critic algorithm.}
    \label{fig:avg_rewards}
\end{figure}

Figure \ref{fig:rate_jock_reneg} compares the reneging and jockeying rates for two queues under Markov against the actor-critic. Reneging rises quickly and stabilizes high while jockeying peaks when the queues are shorter then decays to near-zero. 
Figure \ref{fig:rates_info_src} provides then a queue specific comparison of the impatience. 
The raw Markov-state estimator is noisier and yields higher steady rates, while the actor-critic is smoother, shows decaying jockeying activity with increasing queue lengths. Generally, the actor-critic slightly reduces reneging and jockeying at finite backlogs but matches the Markov model asymptotically.
A persistent multiplicative gap in jockeying rates between the analytic Markov feed and the learned actor–critic is here observed. This difference reflects finite-regime tradeoffs — the actor–critic is optimized for empirical reward and robustness, while the Markov rule is optimized for immediate expected remaining time.
Figures \ref{fig:avg_rewards} recorded the Actor/critic training losses for all episode. All losses rapidly decrease during early training and stabilize by $\sim 20-40$ episodes, consistent with two-timescale actor–critic learning dynamics.
These results illustrate that information format matters at practical or finite queue sizes but the asymptotic behavior is essentially information-agnostic. 
This supports the analytical claim that value-of-information is most relevant in finite regimes, not in the asymptotic limit.

\section{Conclusion and Outlook}
\label{sec:conlusion}
The practical value of system-status signaling for impatient tenants is understudied despite strong industry momentum toward AI-assisted, edge-native control. While richer, more frequent updates can help avoid bad jockeying/reneging decisions, they also consume control-plane resources and can provoke wasteful migrations or stale views (\cite{kiggundu2025nfvsdn}). Our analysis shows that under total-time patience and heterogeneous rates the asymptotic decision outcome is information-agnostic, yet at realistic backlogs the type and cadence of updates materially affect delay, reneging and transient switching. Bridging theory to deployment therefore requires compact testbeds that measure signaling cost (bytes/sec), freshness (age/error), action latency, and end-to-end key performance indicators so one can quantify value-of-information per byte and identify regimes where learning justifies extra telemetry.

\section{Acknowledgment}
This work is supported in fully by the German Federal Ministry of Research, Technology and Space (BMFTR) within the \textit{Open6GHub+} project under grant numbers \textit{16KIS2402K}.

\begin{appendices}

\section*{Appendix}
\label{sec:appendix}
\begin{proof} 
\textbf{II.1: }Adopting \eqref{eqn:waiting_times}, we define $W_1(n)$ as the expected sojourn time in queue 1 with $n$ jobs ahead as the sum of $n$ i.i.d. exponential service times. As $n \to \infty$, $W_1(n)$ grows linearly with $n$, quickly exceeding any fixed patience $T$. 
Similarly, $W_2(m')$ is the expected sojourn time in queue 2 if the customer switches when queue 2 has $m'$ jobs. 
The total sojourn if the customer switches as is then defined using \eqref{eqn:total_jockey_wait}.
    \begin{equation}
        W_{\rm total}=W_1(n)+W_2(m'),
        \label{eqn:total_jockey_wait}
    \end{equation}    
\textit{where for our fixed backlog assumption we take $m'=m$ finite. }

Under the above assumptions, as $n\to\infty$, we show that $\Pr(\text{reneging})=\Pr(W_{\rm total}>T)\to 1$ while the $\Pr(\text{successful jockey})\to 0.$ 

\textbf{Lemma: Chernoff / Cram\'er bound for Exponentials:}
Let $(X_1,X_2,\dots)$ be i.i.d. random variables and $(S_n=\sum_{i=1}^n X_i)$. For any $(x>0)$,
\begin{equation}
\small
    \Pr\Big(\frac{S_n}{n}\ge x\Big)\le \exp\big(-n,I(x)\big), 
    \Pr\Big(\frac{S_n}{n}\le x\Big)\le \exp\big(-n,I(x)\big),
\end{equation}
where the rate function is $I(x)=x-1-\ln x,(x>0)$, (which is nonnegative and $(I(1)=0)$) \cite{Dembo1998,BoucheronLugosiMassart2013}. 

\noindent From this Lemma, the empirical mean of exponential service times satisfies a Cramér-type bound, so the tail decay is exponential with rate ($I(\cdot)$). 
This convergence is exponentially fast under the exponential service assumption. 

Applying the \textbf{Law of Large Numbers (LLN)} to the sum of $n$ i.i.d. $Exp(\mu_1)$ random variables implies the empirical mean converges  to the expected value:
\begin{equation}
     \small
\frac{W_1(n)}{n} = \frac{1}{n}\sum_{i=1}^n X_i \xrightarrow{\text{a.s.}}\mathbf{E}[X_i] = \frac{1}{\mu_1}>0
\end{equation}
Therefore, $W_1(n)\to\infty$ almost surely (and with exponentially small probabilities of large downward deviations). Thus for any fixed constant $T>0$,
\begin{equation}
    \small
    \Pr\{W_1(n)\le T\}\xrightarrow{n\to\infty}0 
\end{equation}
at an exponential rate.
\noindent Then, because $W_{\rm total}\ge W_1(n)$, the probability that the request does not abandon the queue is defined by:
\begin{equation}
    \small
    \Pr(\text{no reneging})=\Pr(W_{\rm total}\le T)\le \Pr(W_1(n)\le T)\xrightarrow{n\to\infty}0. 
    \label{eqn:no_renege}
\end{equation}

\noindent Therefore $\Pr(\text{reneging})=\Pr(W_{\rm total}>T)\to1$. This proves that reneging from the queue becomes the only option. 

Let $A$ be the event that a request switches to an alternative queue and is served, then
\begin{equation}
    A \subseteq \{W_{\rm total}\le T\}, 
\end{equation}
because being served before abandoning implies the total sojourn did not exceed patience $T$. Therefore
\begin{equation}
    \small
    \Pr(A)\le \Pr(W_{\rm total}\le T). 
\end{equation}

From \eqref{eqn:no_renege}, we have $\Pr(W_{\rm total}\le T)\to0$, so $\Pr(A)\to0$. Hence the probability of successful jockeying vanishes as $n\to\infty$. 
\end{proof}

\begin{proof}
\textbf{II.2: }\textbf{Concentration of true remaining work:}
From the definition \eqref{eqn:waiting_times} of the remaining work when \(n\) jobs are ahead, the Strong Law of Large Numbers (SLLN) implies 
\begin{equation}
    \small
\frac{W_i(n)}{n} \xrightarrow{a.s.} \mathbb{E}[X_{i,1}] = \frac{1}{\mu_i},
\qquad (n\to\infty),
\end{equation}
so \(W_i(n)\) grows linearly in \(n\) with slope \(1/\mu_i\) \cite{Durrett2010}.

\noindent \textbf{Assumption: Sublinear estimator-error (regularity).} Let \(cW_i(n)\) denote an estimator (Markov, learned, etc.) of the true remaining work \(W_i(n)\) when \(n\) jobs are ahead. We assume
\begin{equation}
    \small
    \frac{|cW_i(n)-W_i(n)|}{n}\xrightarrow{p}0 \qquad (n\to\infty),
\end{equation}
i.e. the absolute estimator error grows strictly sub-linearly in the backlog. This is satisfied by typical empirical or parametric mean estimators (errors \(O_p(\sqrt{n})\) or \(O_p(1)\)) and by consistent learned regressors under standard concentration conditions. In all these cases the error is $o(n)$ by classical central limit theory and concentration findings \cite{BoucheronLugosiMassart2013,billingsley1995probability}. The assumption therefore rules out only estimators whose bias or variance grows linearly with the backlog. Therefore,
\begin{equation}
    \small
    \frac{\widehat W_i(n)}{n}=\frac{W_i(n)}{n}+o_p(1)\xrightarrow{\mathbb{P}}\frac{1}{\mu_i}.
\end{equation}
implying the estimates share the same first-order slope $1/\mu_i$ as the true remaining work.

\noindent \textbf{Robustness of pairwise comparison — jockeying: }
We deploy Slutsky’s theorem and the sign test show that the sign of the comparison $\widehat W_1(n)> \widehat W_2(m)$ is the same asymptotically as the sign of $W_1(n)> W_2(m)$.
We can fix finite $m$, then consider the estimator decision event
\begin{equation}
    \small
    E_n:=\{\widehat W_1(n)>\widehat W_2(m)\}.
\end{equation}
Which can be decomposed and rewritten as:
\begin{equation}
    \small
    \begin{split}
        \widehat W_1(n)-\widehat W_2(m)=[W_1(n)-W_2(m)]+[\widehat W_1(n)-W_1(n)] \\ - [\widehat W_2(m)-W_2(m)].
    \end{split}    
\end{equation}
We then divide by $n$, then based on the assumption above $W_2(m)=O(1)$ to obtain
\begin{equation}
    \small
    \frac{\widehat W_1(n)-\widehat W_2(m)}{n}
    =\frac{W_1(n)}{n} - \frac{W_1(m)}{n} + o_p(1)
\end{equation}
Since $m$ is fixed, $W_{2}(m)= O(1)$, $\frac{W_{2}(m)}{n}\to 0.$ Hence
\begin{equation}
    \small
    \frac{\widehat W_1(n)-\widehat W_2(m)}{n} \xrightarrow{\mathbb{P}}\frac{1}{\mu_1}.
\end{equation}

\textbf{Lemma:} If $X_n\xrightarrow{\mathbb{P}}a$ with $a>0$, then $\Pr(X_n>0)\to1$.
\begin{proof}
    pick $\varepsilon=a/2$. Then $\Pr(X_n\le 0)\le\Pr(|X_n-a|>a/2)\to0$.
\end{proof}
\noindent Applying this lemma to $X_n=(\widehat W_1(n)-\widehat W_2(m))/n$. If $\mu_1<\mu_2$ then $1/\mu_1>1/\mu_2$ and in particular $1/\mu_1>0$, so the event $\widehat W_1(n)>\widehat W_2(m)$ occurs with probability $\to1$. If $\mu_1>\mu_2$ the limit is negative and the probability tends to $0$. Thus the sign of the estimator comparison is determined, to first order, by $1/\mu_1$.
If $\mu_1<\mu_2$ then $1/\mu_1>1/\mu_2$ and therefore both the estimator and the true comparison declare a beneficial switch with probability tending to $1$; if $\mu_1>\mu_2$ the probability tends to $0$.

\textbf{Inevitability of reneging under total time patience:}
Let $W_{\rm total}$ denote the customer’s total sojourn (including any time spent after switching). Clearly $W_{\rm total}\ge W_1(n)$.
Since $W_1(n)/n\to1/\mu_1>0$, we have $W_1(n)\to\infty$ a.s., hence for any fixed patience $T$
\begin{multline}
    \small
  \Pr(\text{served before }T)\le\Pr(W_{\rm total}\le T) \\
  \le\Pr(W_1(n)\le T)\xrightarrow{n\to\infty}0
\end{multline} 
Thus $\Pr(\text{reneging})\to1$. Because any successful jockey (switch and be served before $T_{local}$) is contained in $\{W_{\rm total}\le T_{local}\}$, the probability of successful jockeying tends to $0$. 
\end{proof}


\end{appendices}

\begin{acronym}[HRTEM]
  \acro{QoS}{Quality of Service}
  \acro{QoE}{Quality of Experience}
  \acro{AnLF}{Analytics logical function}
  \acro{AI}{Artificial Intelligence}
  \acro{NSSF}{Network Slice Selection Function}
  \acro{S-NSSAI}{Single – Network Slice Selection Assistance Information}
  \acro{NSSI}{Network Slice Subnet Instance}
  \acro{NSI ID}{Network Slice Instance ID}
  \acro{NWDAF}{Network Data Analytics Function}
  \acro{SMF}{Session Management Function}
  \acro{SDN}{Software Defined Networks}
  \acro{3GPP}{Third Generation Partnership Project}
  \acro{FCFS}{First Come First Serve}
  \acro{LCFS}{Last Come First Serve}
  \acro{V2X}{Vehicle to Everything}
  \acro{SBA}{Service based Architectures}
  \acro{5G}{Fifth Generation}
  \acro{6G}{Sixth Generation}
  \acro{UEs}{User Equipment}
  \acro{MDP}{Markov Decision Processes}
  \acro{M/M/C}{Markovian/ Markovian/ number of queues}
  \acro{CTMC}{Continuous Time Markov Chain}
  \acro{M/G/C}{Markovian/General/ number of queues}
  \acro{MEC}{Multi-access Edge Computing}
  \acro{RAN}{Radio Access Network}
  \acro{IoT}{Internet of Things}
  \acro{O-RAN}{Open Radio Access Network}
  \acro{DU}{Distributed Unit}
  \acro{RU}{Radio Unit}
  \acro{CU}{Centralized Unit}
  \acro{SIRO}{Serve In Random Order}
  \acro{CTMC}{Continuous Time Markov Chain}
  \acro{SIRO}{Serve In Random Order}
  \acro{CDF}{Cumulative Distribution Function}
  \acro{S-NSSAIs}{Single – Network Slice Selection Assistance Information}
  \acro{M/M/1}{Markovian/Markovian/1}
  \acro{NSSAI}{Network Slice Selection Assistance Information}
  \acro{LDP}{ Large Deviations Principle}
  \acro{PMF}{Probability Mass Function}
  \acro{IID}{Identical and Independently Distributed}
  
\end{acronym}
\scriptsize
\printbibliography
\end{document}